# Unsupervised clustering and classification of upper limb EMG signals during functional movements: a data-driven

Salazar Álvarez L. F. [1][0009-0004-5800-5039], Escobar-Saltarén D. [1][0009-0009-0708-4087], Salazar Sánchez M. B.[1][0000-0002-6137-6357], and Henao-Aguirre S. C. [2][0000-0003-2146-9230]

[1] In2Lab, Engineer Faculty, Universidad de Antioquia. Medellín, Calle 70 No. 52-21, A.A. 1226

[2] School of Engineering and Sciencies, Tecnológico de Monterrey. Guadalajara, México

**Abstract.** This study presents a comprehensive approach for the clustering and classification of upper-limb surface electromyography (sEMG) signals during functional reach and grasp movements. The methodology was applied to the NINAPRO DB4 dataset, which provides multichannel EMG recordings of 52 gestures. A four-stage pipeline was designed, including signal preprocessing, feature extraction, gesture selection via hierarchical clustering, and comparative model evaluation. Preprocessing involved a fourth-order low-pass filter (0.6 Hz) and Hilbert envelope transformation, effectively reducing noise and enhancing signal clarity. Feature extraction yielded 26 temporal and frequency-domain metrics, which were later refined using visual analysis, mutual information, principal component analysis, and decision tree importance scores. A final subset of five key features was selected for classification tasks. Gesture selection was performed through hierarchical clustering using Mahalanobis distance, resulting in six representative movements that balanced biomechanical diversity and computational efficiency. A 200 ms window was identified as optimal for temporal segmentation based on stability and physiological plausibility. Classifier models were evaluated in two stages. Automated comparison using PyCaret identified Extra Trees (ET) and Artificial Neural Networks (ANN) as top performers. Subsequent independent training confirmed their stability and generalization capacity, with ANN showing progressive learning and ET maintaining robust, consistent results. The findings support the implementation of adaptive, low-latency control strategies for myoelectric prostheses and provide a scalable pipeline for future real-time applications.

## 1 Introduction

Upper-limb amputations constitute a public-health problem, accounting for 4.5 % of all reported amputations and exerting a disproportionate impact on the functional autonomy and quality of life of those affected [1]. The inability to perform everyday tasks as simple as reaching for objects has physical, emotional, psychological and social repercussions for the individual. Worldwide prevalence is estimated at 200–300 cases per 100 000 inhabitants, with a relevant share involving the upper extremities. In Colombia this figure warrants a distinct analysis because of the complex interplay of sociopolitical violence, occupational accidents and road traffic incidents—factors that raise the incidence of limb loss.

In recent years, engineering advances have driven the design of myoelectric prostheses with multiple degrees of freedom, integrated sensors and increasingly lightweight interfaces. Yet the continuing challenge is to achieve intuitive control that naturally replicates the complexity of human movement while ensuring comfort and ease of use. Even with sophisticated solutions, contemporary abandonment rates remain close to 35 % owing to issues such as weight, limited wrist dexterity, insufficient grip strength, heat build-up, high cost and—above all—the difficulty users face in perceiving the device as an extension of themselves [2].

Modern approaches process EMG in the time and frequency domains, using statistical features to capture different aspects of muscle activation. Feature selection—through principal-component analysis, decision trees or mutual information—reduces dimensionality and enhances the interpretability of classification models that discriminate movements. These models learn the nonlinear relationship between the feature vector and specific gestures while striving to generalize to new instances and mitigate overfitting. To evaluate such methods, researchers rely on public datasets. The NINAPRO DB4 (Non-Invasive Adaptive Hand Prosthetics) database used in this study provides multichannel sEMG recordings from 12 electrodes sampled at 2 kHz in healthy subjects performing 50 static and functional gestures [3]. Six repetitions per gesture, demographic metadata and carefully curated labels make it a reference standard for training and validating movement-recognition algorithms. Framed within the research line of biological-signal characterization, the present study focuses on reach and grasp movements—essential for fine manipulation and, consequently, patient independence. Our objectives are, first, to cluster grips based on sEMG dynamics, applying statistical criteria to select the most distinguishable classes via Mahalanobis distance; and second, to implement robust, adaptive and interpretable classification models that can underpin more precise and intuitive control systems.

Ultimately, improving motor-intent detection will not only enhance prosthetic functionality but also reduce abandonment rates, foster motor rehabilitation and restore the psychosocial well-being of individuals with upper-limb amputations.

## 2 Data source

Using NINAPRO as the information source for this project, it is relevant to note that the complete database is divided into 10 subsets (DB1–DB10), each with its own acquisition protocol and number of recorded movements. This study focuses on DB4, which is designed for classification and regression tasks in gesture recognition. The selected subset contains recordings from 10 healthy volunteers (6 men and 4 women, with an average age of about 29 years) as shown in Fig. 1.

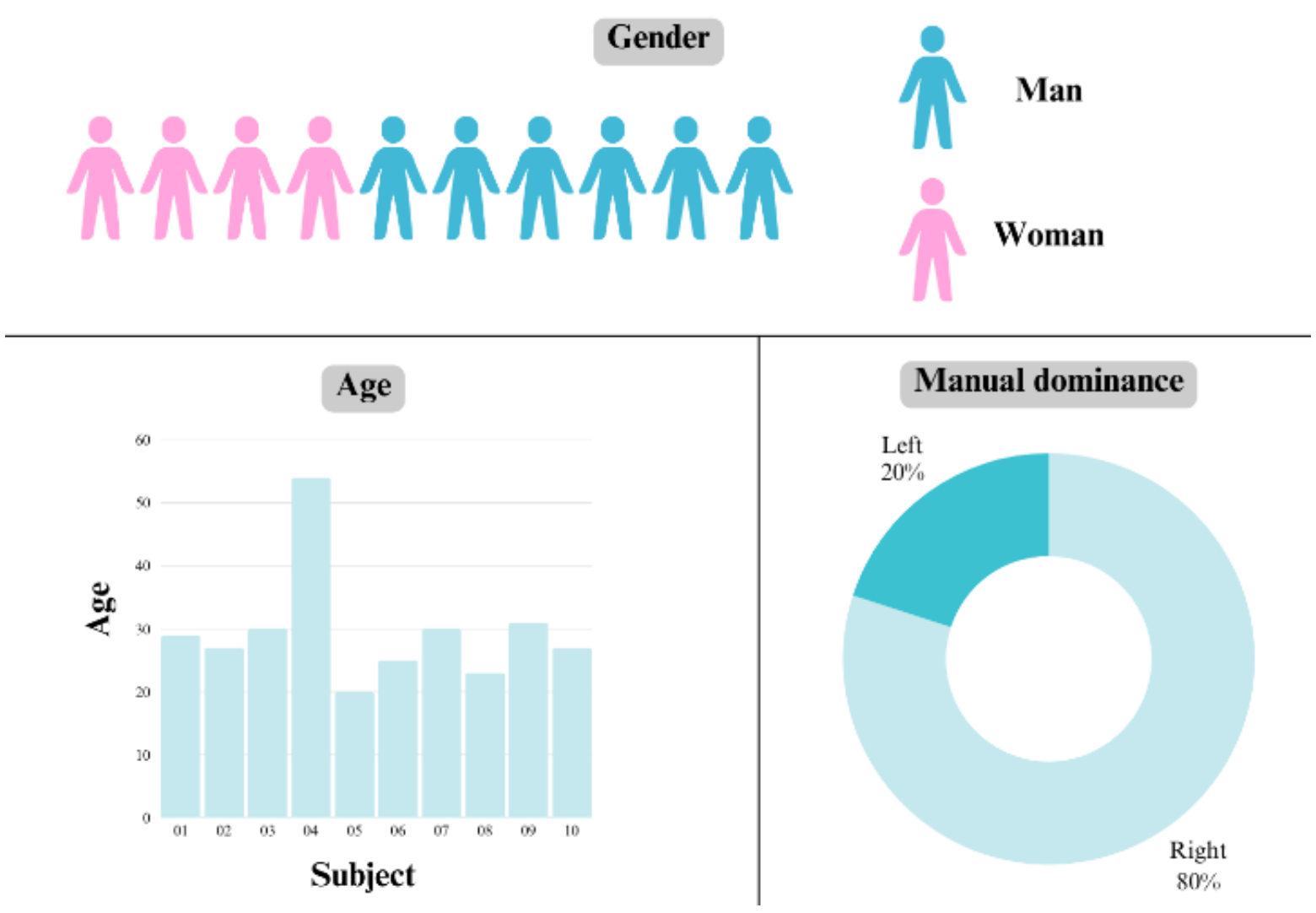

**Fig. 1.** Database - DB 4 demographic information

Their sEMG signals were captured with 12 Cometa electrodes placed around the forearm—from the distal biceps' region to the extensor and flexor musculature of the forearm (Fig. 2)—to cover the main muscle groups involved in reaching, gripping, and pronation-supination. Each channel was digitized at 2 kHz, enabling analyses in both the time and frequency domains.

The protocol comprises 52 distinct gestures, shown in Fig. 3. Each gesture is repeated six times per subject: 5 seconds of sustained contraction followed by roughly 3 seconds of rest. This design strikes a balance between minimal fatigue and enough training samples. The original files contain time-synchronized variables, making it straightforward to align muscle signals with protocol events—an essential feature for developing prosthetic-control algorithms and functional-rehabilitation tools.

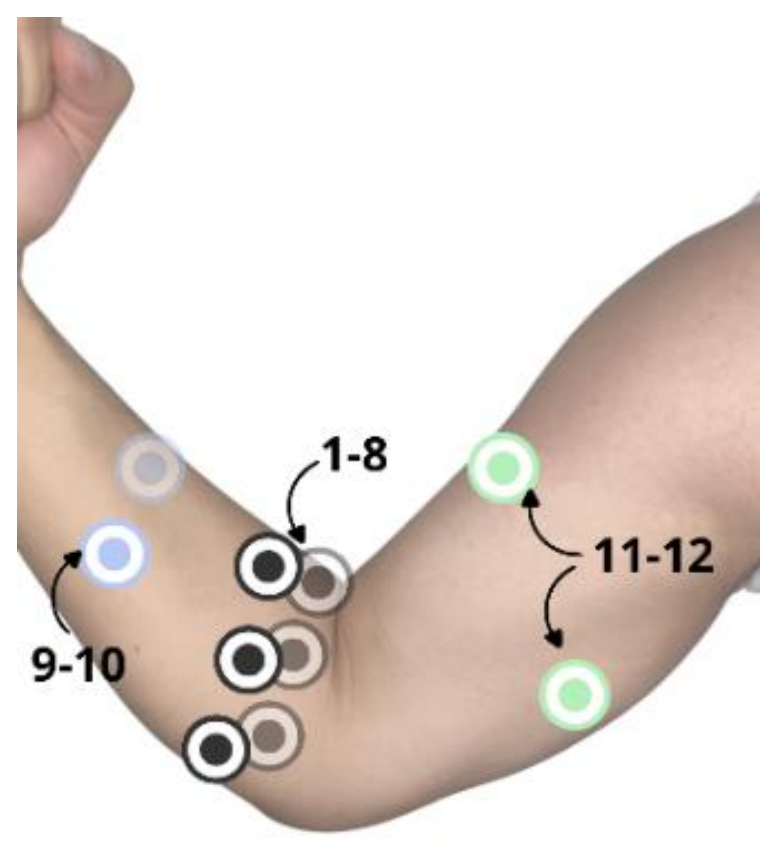

**Fig. 2.** Anatomical disposition of the electrodes

The present article presents the insights gained from characterizing sEMG recordings obtained during functional upper-limb movements, using the public **NINAPRO DB4** database. The methodology is organized into four main stages: (A) signal preprocessing and transformation, (B) feature extraction, (C) movement selection by cluster analysis, and (D) classifier selection and validation. The overarching goal is to identify muscle-activation patterns with sufficient discriminative power to feed movement-classification algorithms for prosthetic control.

| A | | B | | | | C | | | | REST |
|---|---|---|---|---|---|---|---|---|---|---|
| 50 | A1 | 1 | B1 | 13 | B13 | 18 | C1 | 30 | C13 | |
| 51 | A2 | 2 | B2 | 14 | B14 | 19 | C2 | 31 | C14 | |
| 52 | A3 | 3 | B3 | 15 | B15 | 20 | C3 | 32 | C15 | |
| 53 | A4 | 4 | B4 | 16 | B16 | 21 | C4 | 33 | C16 | |
| 54 | A5 | 5 | B5 | 17 | B17 | 22 | C5 | 34 | C17 | |
| 55 | A6 | 6 | B6 | | | 23 | C6 | 35 | C18 | |
| 56 | A7 | 7 | B7 | | | 24 | C7 | 36 | C19 | |
| 57 | A8 | 8 | B8 | | | 25 | C8 | 37 | C20 | |
| 58 | A9 | 9 | B9 | | | 26 | C9 | 38 | C21 | |
| 59 | A10 | 10 | B10 | | | 27 | C10 | 39 | C22 | |
| 60 | A11 | 11 | B11 | | | 28 | C11 | 40 | C23 | |
| 61 | A12 | 12 | B12 | | | 29 | C12 | | | |

**Fig. 3.** Movements recorded in DB4 and their respective labeling

# 3 Methodology

## 3.1 Signal preprocessing and transformation

After downloading the raw DB4 files, a pipeline was applied to attenuate noise generated by motion artefacts, highlight active contraction phases, and ensure inter-subject comparability:

a) Electrical-noise filtering: mains interference (60 Hz) was removed.
b) Butterworth low-pass filter: a fourth-order digital filter with a 0.6 Hz cut-off frequency smoothed abrupt variations without affecting the low-frequency content relevant to muscle activation.
c) Envelope calculation (Hilbert): the Hilbert transform was used to obtain the analytic envelope of each channel, providing a clearer and more continuous representation of the signal.
d) Subject- and channel-wise normalization: each signal was scaled to reduce intrinsic variability between individuals.

## 3.2 Feature extraction

Statistical features were extracted from the processed signals in both the time and frequency domains. Selected for their balance of low computational cost and high discriminative capacity reported in the literature, these metrics included mean absolute value (MAV), root mean square (RMS), integrated absolute value (IAV), and the wavelet-based mDWT. In addition, the standard deviation of each metric was computed, yielding an initial set of 26 attributes per signal segment. Some of these features are listed in Table 1 below.

Table 1. Extracted features categorized by domain and functional grouping

| ***Feature*** | ***Domain*** | ***Group*** |
|---|---|---|
| RMS (Root Mean Square) | Time Domain | Energy/Amplitude |
| MAV (Mean Absolute Value) | | |
| IAV (Integrated Absolute Value) | | |
| WL (Waveform Length) | | Slope/Change |
| MAVS (Mean Absolute Value Slope) | | Signal Pattern |
| ZC (Zero Crossing) | | |
| SSC (Slope Sign Changes) | | Variability/Dispersion |
| VAR (Variance) | | |
| CoV (Coefficient of Variation) | | |
| Autorregresive | | Statistical |
| Kurt | | |
| MNP (Mean Power) | Frequency Domain | Signal Pattern |
| MNF (Mean Frequency) | | |
| mDWT (Marginal Discrete Wavelet) | | |

### 3.3 Movement selection by cluster analysis

It is important to highlight the selection of a representative, non-redundant subset of functional movements as a key phase in the development of this project. As several authors note [2], reducing overlap between classes and discarding irrelevant gestures improves a model's generalization capability, shortens training times, and prevents inter-class ambiguities that often translate into control errors during prosthesis use.

After the feature extraction, an exploratory inspection of the 52 Ninapro gestures was undertaken. Gesture similarity was assessed via hierarchical cluster analysis, with the Mahalanobis distance chosen as the proximity metric. Unlike Euclidean distance, Mahalanobis accounts for the variance–covariance structure of the features, penalizing highly correlated dimensions and thus providing a more robust criterion in the presence of internal redundancies [4]. The clustering strategy employed complete linkage, which defines inter-cluster distance as the largest Mahalanobis distance between any pair of members from different clusters. This method tends to yield compact, well-separated groups, desirable for minimizing overlap in feature space [4].

Results were visualized with a dendrogram, revealing six major clusters that satisfied two fundamental principles:

a) Maximum intra-cluster homogeneity, evidenced by reduced variances within feature vectors

b) Mínimum inter-cluster similarity

Within each macro-cluster, the most representative movement was selected—defined as the gesture with the highest energy (based on RMS analysis) and the greatest potential for replicability in future studies.

### 3.4 Temporal segmentation of the signal

Once filtering, per-channel normalization and envelope calculation were completed, the sEMG signal underwent temporal segmentation—a critical phase for ensuring that the extracted features capture the true dynamics of each gesture without introducing biases from fatigue or overlap between consecutive movements. Following guidelines identified in the literature review, three non-overlapping rectangular window lengths were chosen—100 ms, 200 ms and 300 ms—which also exhibit an inverse relationship with the system's computational demands. Windows were defined on the analytic envelope of each channel so that every segment contained an isolated portion of myoelectric activity associated either with the sustained-contraction phase or the corresponding rest period. To assess the reliability and replicability of the metrics, the 26 previously described features were extracted from each individual window, after which graphical representations—boxplots and violin plots—were constructed.

### 3.5 Feature Selection

Feature selection has a direct impact on model accuracy, generalization capacity, and computational efficiency. A hybrid scheme was designed that combines exploratory inspection, statistical indicators, and model-derived criteria to ensure that the final feature set captures muscle activity without unnecessary redundancies.

a) Visual exploratory análisis. Boxplots and violin plots were generated for each of the 26 features across the six selected gestures, making it easier to detect extreme biases that could undermine classifier robustness and to identify inter-class overlap, thus visually gauging the discriminative relevance of each feature.

b) Statistical dependence via Mutual Information (MI). MI was computed between every feature and the gesture label and then normalized to the 0–1 range.

c) Multivariate structure with Principal Component Analysis (PCA). PCA was applied to the 12 features retained after the MI filter; components explaining ≥ 85 % of the cumulative variance were kept.

d) Variable importance in preliminary Decision Trees. A classification tree was trained, and the importance score of each feature within the model was extracted for further refinement.

### 3.6 Implementation of classifier models

In the present study, the recognition of the six selected gestures was carried out in two complementary phases, aimed at combining speed and precision, supporting clinical applications where a balance between accuracy and reduced calibration time is required. This multistage approach enabled both a systematic and automated comparison of various algorithms and a detailed evaluation of the top-performing classifiers:

a) PyCaret. To compare multiple classification algorithms under the same validation workflow in an automated manner, the PyCaret library used a free and open-source tool designed to simplify the automation of machine learning pipelines. This allowed for rapid experimentation with low computational cost. A total of 16 algorithms were evaluated, from which the top three were selected for the next phase.

b) Independent Testing. The models with the best preliminary performance were then trained independently using cross-validation and hyperparameter tuning, allowing for a more detailed analysis of their behavior. This process included manual hyperparameter adjustments and fold-by-fold evaluation to assess model stability and generalization capability.

## 4 Results and Discussion

### 4.1 Exploratory analysis and signal preprocessing

In the present study, the recognition of muscle gestures was based on a preliminary analysis of surface electromyography (sEMG) signals extracted from the target database. These signals exhibited typical characteristics of muscle biopotentials, such as low amplitude, the presence of motion artifacts, and electrical interference from the environment. It is important to note that, through visual inspection, high inter-channel variability was observed. This can be attributed to the anatomical placement of electrodes over different muscle groups in the forearm (Figure 2), which exhibit specific activations depending on the performed movement. To address the limitations caused by high inter-channel variability, artifacts, and noise present in the signals, a two-step preprocessing procedure was implemented: (1) a low-pass filter with a cutoff frequency of 0.6 Hz, aimed at attenuating high-frequency components that could hinder the processing; and (2) the application of the Hilbert envelope, which enhances the instantaneous amplitude of the signal and provides a smoothed representation of muscle contraction, thus improving both temporal segmentation and feature extraction.

Fig. 4 shows a comparison between the raw signal (blue) and its corresponding processed signal (orange) for channel 10 of a randomly selected subject. It can be clearly seen how the noise has been effectively suppressed, while the characteristic peaks of muscle activation are preserved, thereby facilitating the physiological interpretation of the signal.

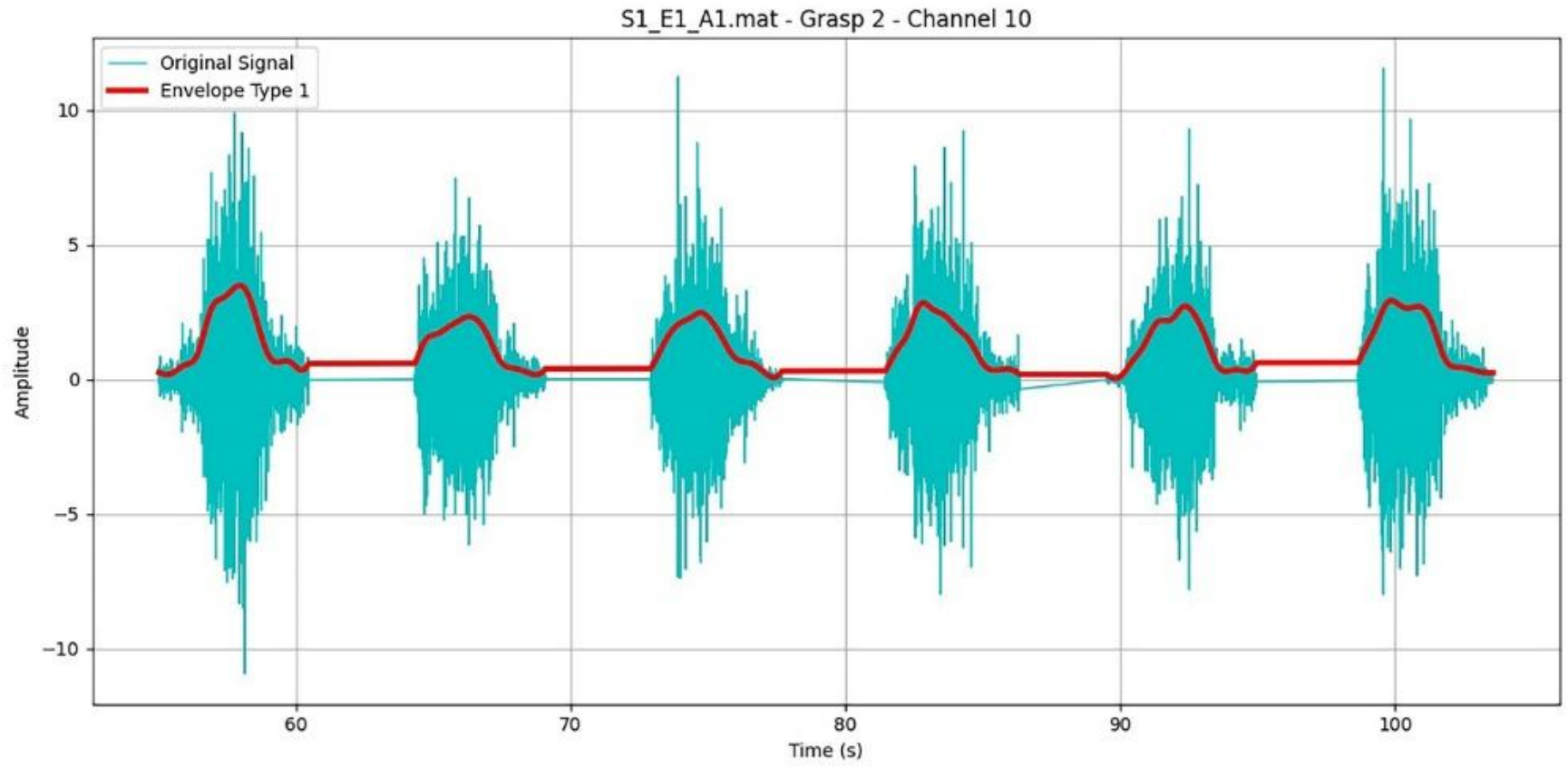

***Fig. 4.*** Raw signal (blue) and processed signal (orange) for channel 10 of a particular subject. Comparison of the original signal with the signal smoothed by low-pass filter and Hilbert envelope, showing noise reduction.

### 4.2 Cluster analysis and movement selection

The selection of functional movements to be included in the classification model is a critical step that enables more reliable performance. This decision was based on the large number of gestures available in the NINAPRO DB4 database (over 50 types), which posed a challenge due to the presence of gesture classes that could be statistically similar or even redundant—introducing ambiguity and affecting the model's generalization capability.

To address this, a complete-linkage hierarchical clustering approach was applied, using Mahalanobis distance as the dissimilarity metric. The result of this analysis was graphically represented through a dendrogram (Fig. 5), which shows the hierarchical grouping of the various muscle gestures. In this visualization, gestures with high statistical similarity are clustered at lower levels, whereas those with greater dissimilarity merge at higher levels. This representation facilitated the identification of naturally differentiable movement subsets (Fig. 6), allowing the selection of those with greater discriminative capacity and high energy levels.

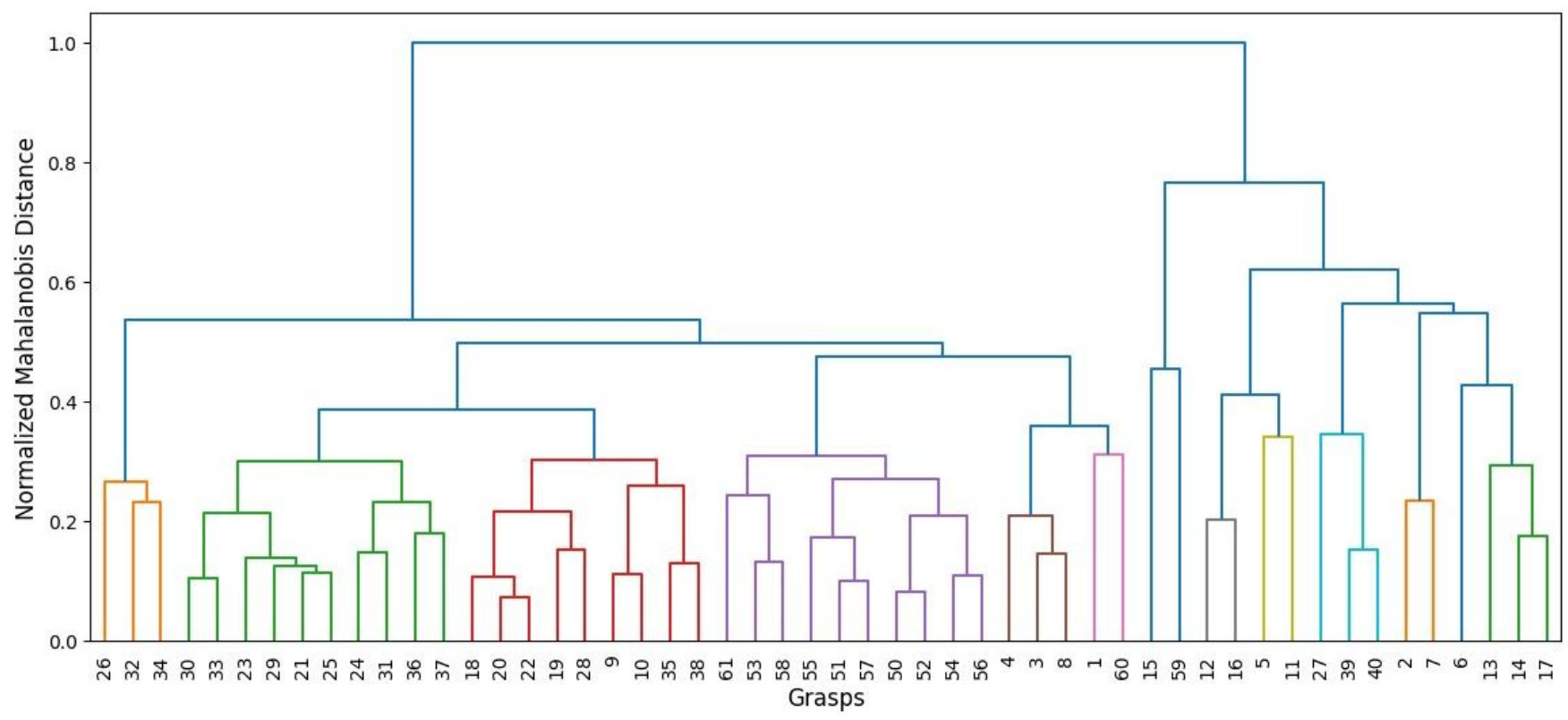

***Fig. 5***. Dendrogram based on Mahalanobis distance, hierarchical grouping of muscular gestures according to their multivariate statistical distance. Allows selection of differentiable movements.

Based on this representation, a final subset consisting of six movements (including the resting state) was selected. These gestures were considered sufficiently representative, distinguishable, and functionally relevant. This selection optimized the balance between biomechanical diversity and simplification of the classification scenario, reducing both the computational load and the complexity of the model.

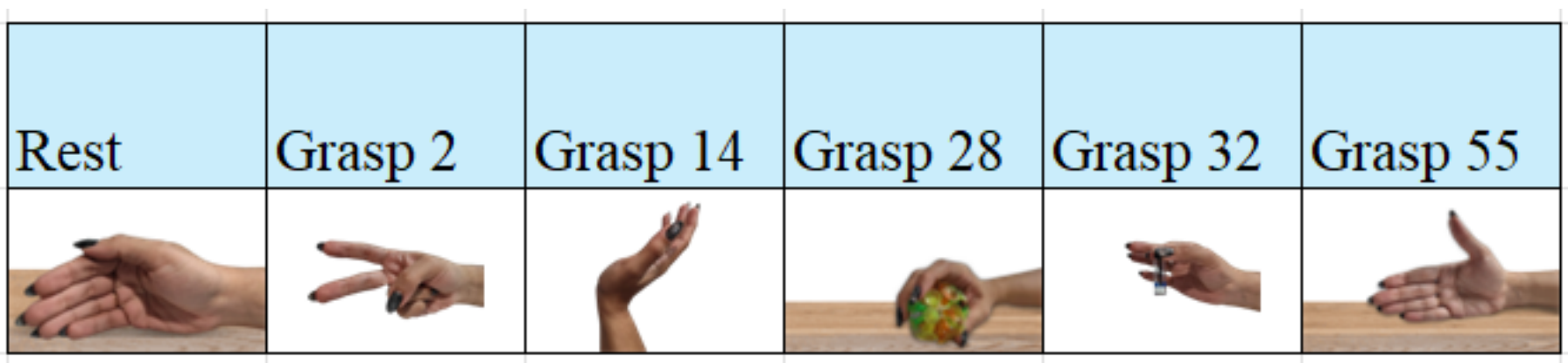

***Fig. 6.*** *Selected gestures based on dendrogram analysis. They represent a functionally diverse and statistically distinguishable subset within the original set of movements from the DB4 dataset.*

### 4.3 Temporary window size evaluation

The selection of the window size for this study was carried out by evaluating three window durations—100 ms, 200 ms, and 300 ms—without overlapping, applying the same set of previously extracted features to each configuration. It was observed that 100 ms windows, although offering high temporal resolution, exhibited high variability in the extracted features, as well as a marked presence of outliers. This instability can hinder class discrimination and lead to an overestimation of noise within the model. In contrast, 300 ms windows significantly smoothed the signal and showed greater consistency across subjects and repetitions, but at the expense of reduced sensitivity to transient muscle activation events, negatively impacting the system's dynamic responsiveness.

On the other hand, the 200 ms window emerged as the most suitable option, providing a balance between resolution and stability. It allowed for the capture of relevant muscle activation patterns with lower susceptibility to noise, resulting in better inter-class discrimination in later processing stages. Moreover, this decision is supported by physiological literature showing that the central nervous system can organize posture-specific muscle activation patterns within time frames close to 200 ms, as reported by studies on motor control and neurophysiological reaction times [5].

### 4.4 Feature selection and debugging

The initial feature extraction phase resulted in a total of 26 features for each window, repetition, movement, and subject. These included metrics from the time domain, frequency domain, and dispersion measures. Subsequently, various techniques were applied to refine and select the most informative features: exploratory visual analysis, mutual information calculation, principal component analysis (PCA), and decision tree training. These tools made it possible to identify features with low discriminative power, redundancy, or statistically negligible contribution to the model.

PCA analysis was particularly useful for examining the internal structure of the dataset. Fig. 7 shows a representation in the space of the first two principal components, where class separation can be observed. Some classes, such as rest, display compact and well-defined groupings, while others, like gestures 28 and 32, show some degree of overlap—although with distinguishable dispersion patterns. This non-homogeneous distribution highlights the inter-subject variability inherent to EMG signals. As a result of the refinement process, all standard deviation features, six additional low-relevance features, and channels 11 and 12 were eliminated due to their consistently low importance across all applied approaches. The final feature set consisted of: IAV, MAV, MAVS, RMS, and mDWT, applied across the 10 most relevant channels.

### 4.5 Comparison of classification models

The classification model comparison stage represented a vital component in the development of this research, allowing for the identification of the most effective approaches for the task of multiclass movement recognition based on sEMG signals. This phase was divided into two main stages: an initial evaluation using the PyCaret library, and a subsequent implementation involving independent hyperparameter tuning and controlled cross-validation. In the first stage, PyCaret facilitated the automated comparison of multiple algorithms under standardized conditions, streamlining the analysis without compromising methodological rigor. Models such as K-Nearest Neighbors (KNN), Support Vector Machines (SVM), logistic regression, decision trees, Extra Trees Classifier (ET), and Artificial Neural Networks (ANN) were evaluated, among others. The metrics used included accuracy, recall, F1-score, precision, and Cohen's Kappa. The results showed that ANN and ET models outperformed the others, achieving accuracy close to 95%, along with high and consistent values across all evaluated metrics (see Table 2). This preliminary performance positioned both models as leading candidates for more in-depth validation.

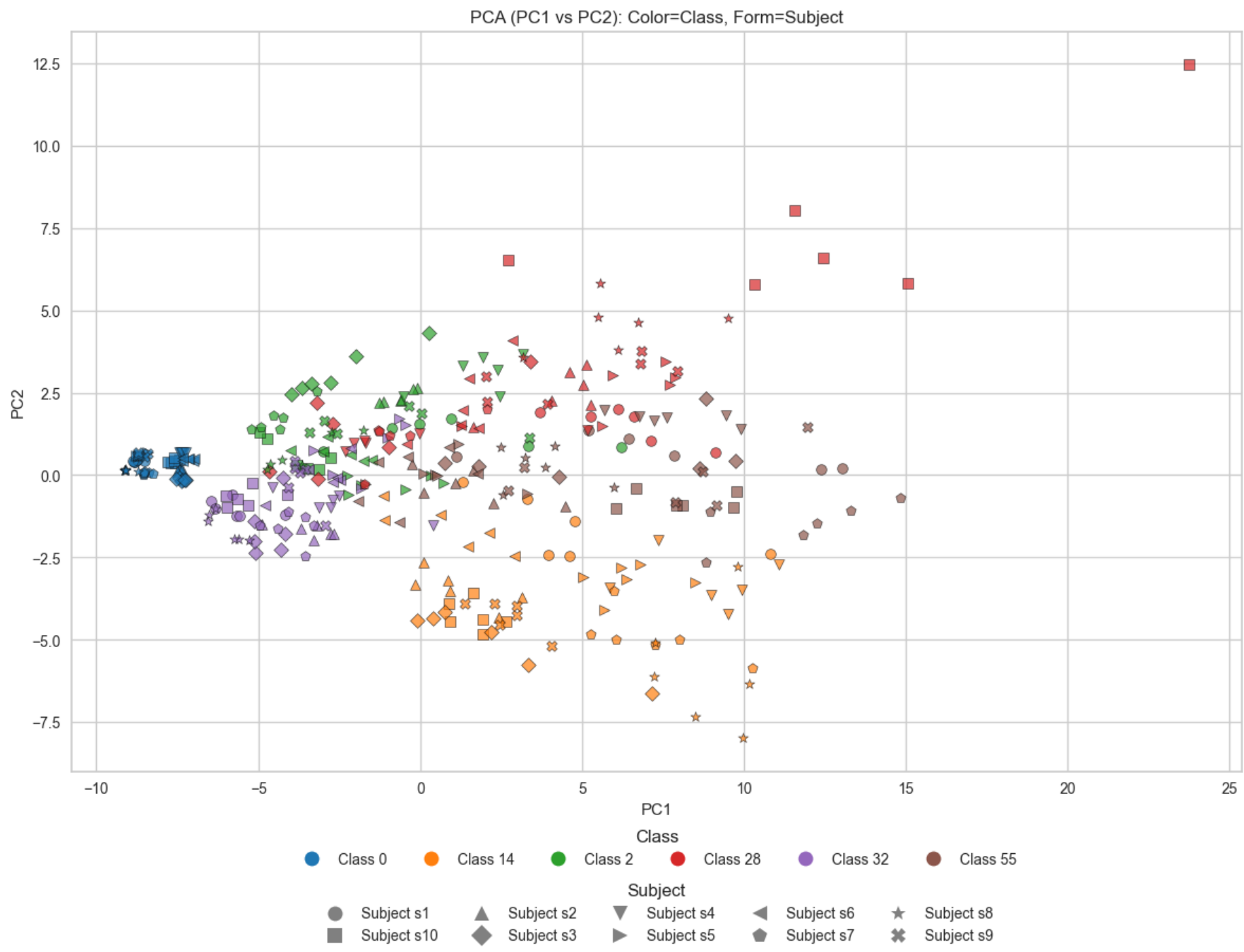

***Fig. 7.*** Representation of EMG signals in the space of the first two principal components (PCA).

Table 2. Performance of RNA and ET models evaluated by PyCaret.

| ***Model*** | ***Accuracy*** | ***Recall*** | ***F1*** | ***Prec.*** | ***Kappa*** |
|---|---|---|---|---|---|
| RNA | 95.2% | 95.2% | 95.7% | 95.2% | 94.3% |
| ET | 94.4% | 94.4% | 94.9% | 94.4% | 93.3% |

In the second phase, the selected models were trained independently, allowing for manual hyperparameter tuning and more detailed cross-validation. This stage aimed to verify the robustness observed in PyCaret within a less automated environment. The results confirmed the stability of both the ANN and ET models, each maintaining high metric values with minimal differences compared to the previous results (see Table 3). At this point, ANN showed a slight improvement in the F1-score, while ET maintained consistent performance, reaffirming its reliability.

Table 3. Performance of RNA and ET models evaluated independently

| ***Model*** | ***Accuracy*** | ***Recall*** | ***F1*** | ***Prec.*** | ***Kappa*** |
|---|---|---|---|---|---|
| RNA | 94.7% | 94.7% | 95% | 94.7% | 93.6% |
| ET | 94.4% | 94.4% | 94.9% | 94.4% | 93.3% |

During the analysis of the learning curves, both models demonstrated stable behavior. In the case of the ET model, the validation curve remained very close to the training curve, indicating a low tendency toward overfitting and good generalization. ANN, on the other hand, showed progressive improvement as the size of the training set increased, suggesting effective and gradual learning—typical of well-configured neural network architectures. This trend was confirmed by the confusion matrices, which reflected a low error rate and few misclassifications between similar gestures.

Additionally, models such as KNN and SVM were also compared, but they showed lower performance, particularly in multiclass tasks with high dimensionality and inter-subject variability. This reinforces the idea that models like ET, due to their ability to handle nonlinear relationships and low risk of overfitting, and ANN, due to their capacity to learn complex representations, are the most suitable for this type of problem.

## 5 Conclusion

This study proposed and validated a comprehensive approach for the characterization, selection, and classification of upper limb sEMG signals during functional reach and grasp movements, within the framework of developing more adaptive control systems for myoelectric prostheses. Using the public NINAPRO DB4 database as the source of information, a step-by-step workflow was designed, covering the transformation of raw signals through to the comparison of classification models.

One of the main contributions of this work lies in the preprocessing strategy applied, based on the use of a fourth-order low-pass filter (0.6 Hz) and the analytic envelope obtained via the Hilbert transform. This combination enabled the attenuation of high-frequency noise and provided a smoother, more continuous representation of muscle activations, improving not only the temporal segmentation of gestures but also the overall quality of the extracted features. This methodology proved particularly useful in prototypical instrumentation contexts, as evidenced by preliminary testing with a custom-designed acquisition armband.

The movement selection stage, using hierarchical cluster analysis with Mahalanobis distance, introduced an objective, data-driven perspective to reduce inter-class ambiguity. This technique allowed the 52 original gestures to be grouped into homogeneous macro-clusters and led to the selection of six representative gestures (including rest), optimizing the balance between biomechanical diversity and computational complexity. This selection was also based on estimated muscle energy (via RMS) and the gestures' potential for replicability.

Regarding temporal segmentation, both empirical and theoretical evidence confirmed that the 200 ms window represents an optimal compromise between temporal resolution and statistical stability of the features. In addition to improving inter-class discrimination, this choice was supported by neurophysiological studies demonstrating the central nervous system's ability to organize specific muscle activation patterns within such time frames—particularly during postural or reactive tasks.

Feature extraction and subsequent refinement were also central to the study. Starting from an initial set of 26 statistical attributes from the time and frequency domains, a hybrid selection strategy was designed using visual criteria (boxplots and violin plots), statistical dependencies (mutual information), multivariate analysis (PCA), and variable importance scores from decision trees. This combination enabled the identification and removal of redundant or low-discriminative features, resulting in a reduced and efficient feature set consisting of IAV, MAV, MAVS, RMS, and mDWT, distributed across 10 channels selected for their high inter-subject relevance.

On the other hand, the staged implementation and comparative evaluation of classification models constituted a significant phase of the project. In the first stage, PyCaret was used to automatically explore the performance of 16 algorithms under standardized conditions. This evaluation revealed that the Extra Trees (ET) and Artificial Neural Networks (ANN) models delivered the best results in terms of accuracy, F1-score, precision, recall, and Cohen's kappa coefficient. In the second stage, both models were independently trained and validated, with manual hyperparameter tuning and fold-based cross-validation. Their stability and generalization capacity were confirmed through learning curves and confusion matrices. ANN showed progressive improvement with increased training data, while ET maintained consistent performance—ideal for real-time implementation.

Alternative models such as KNN and SVM were also evaluated but demonstrated lower performance, particularly in multiclass tasks with high dimensionality and inter-subject variability. Despite these advances, the study also revealed persistent challenges, such as inter-subject variability even after normalization, and the need to incorporate dimensionless features or adaptive scaling techniques. Future lines of research should focus on clinical validation with real users, integration into real-time embedded systems, and the fusion of multimodal signals (EMG, accelerometry, force sensors) to enhance the contextual richness and functionality of myoelectric control systems.

## Acknowledgments

The authors express deep gratitude to the University of Antioquia, this work was supported by CODI projects (PRG2024-77051). We are also due to the Intelligent Information Systems Lab (In2Lab-UdeA) for their technical support and for providing access to the computational resources essential for the analysis of EMG signals. Special appreciation goes to her research colleague for his continuous assistance in data processing and model validation, and to the professor who contributed during the statistical analysis stage, for his guidance in the use of multivariate and machine learning techniques.